\documentclass[lettersize,journal]{IEEEtran}
\usepackage{amsmath,amsfonts}
\usepackage{algorithmic}
\usepackage{graphicx}
\usepackage{algorithm}
\usepackage{array}
\usepackage{tikz}
\usepackage{comment}
\usepackage{multirow}
\usepackage{tablefootnote}
\usepackage[caption=false,font=normalsize,labelfont=sf,textfont=sf]{subfig}
\usepackage{textcomp}
\usepackage{stfloats}
\usepackage{url}
\usepackage{verbatim}
\usepackage{graphicx}
\usepackage{footmisc}
\usepackage[pagebackref=false,breaklinks,colorlinks,bookmarks=false]{hyperref}
\usepackage{colortbl} 
\definecolor{mypink}{cmyk}{0, 0.7808, 0.4429, 0.1412}
\definecolor{mygreen}{rgb}{0.0, 0.7, 0.0}
\definecolor{myblue}{rgb}{0.0, 0.72, 0.92}
\definecolor{mygray}{gray}{0.6}
\definecolor{mygray-bg}{gray}{0.9}

\newcommand{\ie}{\textit{i}.\textit{e}.}
\newcommand{\eg}{\textit{e}.\textit{g}.}
\newcommand{\cf}{\textit{cf.}}
\newcommand{\etal}{\textit{et}.\textit{al}.}

\newcommand{\vs}{\textit{vs}.}
\usepackage{amssymb}

\begin{document}
\title{Label Semantic Knowledge Distillation for Unbiased Scene Graph Generation}

\author{Lin Li,
        Long Chen$^*$,
        Hanrong Shi,
        Wenxiao Wang,
        Jian Shao,
        Yi Yang,
        Jun Xiao
\thanks{$^*$Long Chen is the corresponding author.}
\thanks{L. Li, H. Shi, W. Wang, J. Shao, Y. Yang and J. Xiao are with Zhejiang University, Hangzhou, 310027, China. Email: mukti@zju.edu.cn, hanrong@zju.edu.cn, wenxiaowang@zju.edu.cn, jshao@cs.zju.edu.cn, yangyics@zju.edu.cn, junx@cs.zju.edu.cn.}
\thanks{L. Chen is with Columbia University, New York, 10027, USA. Email: zjuchenlong@gmail.com.}
}

\markboth{Journal of \LaTeX\ Class Files,~Vol.~xx, No.~xx, August~2022}%
{Shell \MakeLowercase{\textit{et al.}}: Label Semantic Knowledge Distillation for Unbiased Scene Graph Generation}



\maketitle

\begin{abstract}
The Scene Graph Generation (SGG) task aims to detect all the objects and their pairwise visual relationships in a given image. Although SGG has achieved remarkable progress over the last few years, almost all existing SGG models follow the same training paradigm: they treat both object and predicate classification in SGG as a single-label classification problem, and the ground-truths are one-hot target labels. However, this prevalent training paradigm has overlooked two characteristics of current SGG datasets: 1) For positive samples, some specific subject-object instances may have multiple reasonable predicates. 2) For negative samples, there are numerous missing annotations. Regardless of the two characteristics, SGG models are easy to be confused and make wrong predictions. To this end, we propose a novel model-agnostic Label Semantic Knowledge Distillation (LS-KD) for unbiased SGG. Specifically, LS-KD dynamically generates a ``soft" label for each subject-object instance by fusing a predicted Label Semantic Distribution (LSD) with its original one-hot target label. LSD reflects the correlations between this instance and multiple predicate categories. Meanwhile, we propose two different strategies to predict LSD: iterative self-KD and synchronous self-KD. Extensive ablations and results on three SGG tasks have attested to the superiority and generality of our proposed LS-KD, which can consistently achieve decent trade-off performance between different predicate categories.
\end{abstract}

\begin{IEEEkeywords}
Unbiased Scene Graph Generation, Knowledge Distillation
\end{IEEEkeywords}

\section{Introduction}
\label{sec:1}
\begin{figure*}[t]
  \centering
  \includegraphics[width=0.8\linewidth]{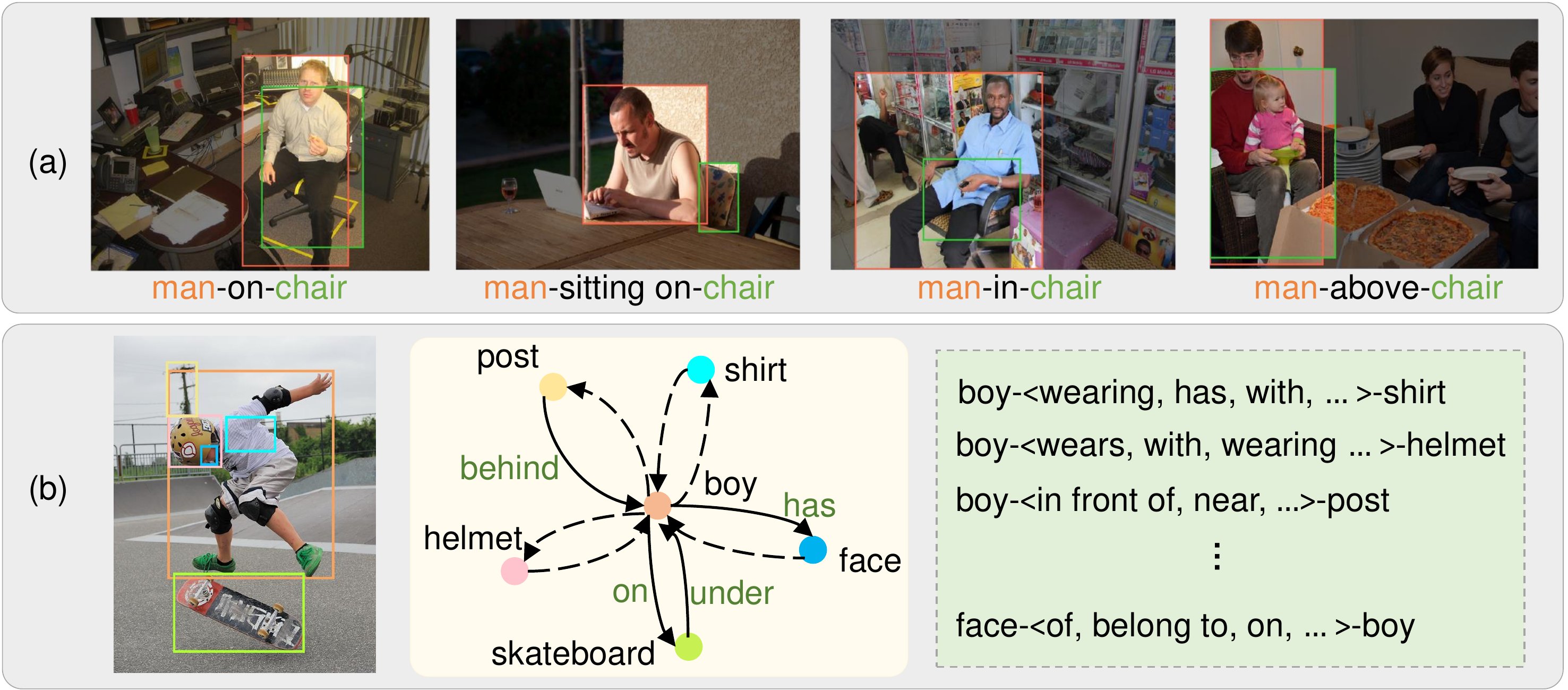}
  \caption{Illustration of two overlooked characteristics in SGG benchmarks~(take VG as an example). a) All listed \texttt{man}-\texttt{chair} pairs with similar visual appearance are assigned various predicates with semantic overlap and overlay. These predicates are also reasonable for other \texttt{man}-\texttt{chair} pairs. b) Images always have missing-annotated issues, and the missing-annotated triplets with one/many ``reasonable" predicate(s) are directly treated as absolute \texttt{background}. Solid lines are annotated relations, dotted lines are missing-annotated, and green boxes are latent predicates for missing-annotated triplets.}
  \label{fig:intro}
\end{figure*}

Scene Graph Generation (SGG), \ie, detecting all the objects (including both localization and classification) and classifying their pairwise visual relationships in a given image, is a challenging and fundamental scene understanding task~\cite{xu2017scene}. Typically, a scene graph can be formulated as a series of \texttt{subject}-\texttt{predicate}-\texttt{object} triplets (\eg, ``\texttt{man}-\texttt{on}-\texttt{chair}" in Fig.~\ref{fig:intro}(a)). Due to its symbolic representation nature, SGG has played an important role in many downstream tasks, such as image captioning~\cite{yang2019auto,li2019know,gu2019unpaired,yu2021dual,mao2022rethinking}, image retrieval~\cite{johnson2015image,bai2021unsupervised}, visual question answering~\cite{shi2019explainable,zhang2019frame,chen2020counterfactual}, and so on.  Meanwhile, with the release of several large-scale SGG datasets (\eg, Visual Genome (VG)~\cite{krishna2017visual}) and the maturity of advanced object detectors~\cite{ren2015faster}, SGG has raised unprecedented attention over the recent few years, and hundreds of SGG models have been proposed~\cite{ li2022ppdl,9829320,dong2022stacked,goel2022not}.

However, one notorious challenge in existing SGG benchmarks is the highly-skewed long-tailed relation annotations, \ie, the head predicate categories\footnote{For brevity, we directly use ``tail", ``body", and ``head" categories to represent the predicate categories in the tail, body, and head parts of the number distributions of different ground-truth predicate annotations in SGG datasets, respectively. Besides, we use ``positive" and ``negative" samples to denote annotated ground-truth visual relation triplets with foreground and background predicate categories, respectively. \label{footnote:1}} have much more annotated ground-truth samples (or relation triplets) than the tail ones. Thus, a recent surge of interest in SGG is developing \emph{unbiased} scene graph generation models, which can achieve more balanced performance on all predicate categories (\ie, head, body, and tail categories). Currently, existing state-of-the-art unbiased SGG models can be coarsely grouped into four categories: 1) \emph{Class-aware Re-sampling}: They try to sample more tail category training samples to balance the data distribution~\cite{li2021bipartite}. 2) \emph{Loss Re-weighting}: They aim to reduce the contributions of the head categories training samples in the loss calculation by designing different weighting strategies~\cite{lin2017focal,yan2020pcpl,knyazev2020graph}. 3) \emph{Biased-model-based}: They disentangle unbiased representations from some biased trained  models~\cite{tang2020unbiased,chiou2021recovering,yu2021cogtree,guo2021general}. 4) \emph{Noisy Label Learning}: They first detect noisy annotated samples and then reassign them to more high-quality predicate labels~\cite{li2022devil,li2022nicest}. Although different advanced techniques have been proposed in each category, almost all existing unbiased SGG models follow the same training paradigm: \emph{regarding the object and predicate classification as a \textbf{single-label} classification problem, and using the cross-entropy loss (between predicted distributions and one-hot target labels) as the training objective}.

Although today's unbiased SGG models have achieved significant progress on some challenging SGG benchmarks (\eg, VG), it is worth noting that this training paradigm has overlooked two characteristics of existing SGG datasets: 1) For the positive samples\footref{footnote:1}, some specific subject-object instances may have multiple reasonable predicates, \ie, the semantics of different categories are not completely independent. For example in Fig.~\ref{fig:intro}(a), predicates ``\texttt{on}", ``\texttt{sitting on}", ``\texttt{in}", and ``\texttt{above}" are all reasonable predicates in these ``\texttt{man}-\texttt{chair}" pairs. 2) For the negative samples, there are numerous missing annotations~\cite{lu2016visual}, and all these negative samples are treated as absolute \texttt{background} category in the training process. For instance, ``\texttt{boy}-\texttt{shirt}", ``\texttt{boy}-\texttt{helmet}" in Fig.~\ref{fig:intro}(b) are all missing-annotated subject-object instances. Similarly, some ``negative" samples may even have multiple reasonable predicates (\eg, ``\texttt{wearing}" and ``\texttt{has}" for ``\texttt{boy}-\texttt{shirt}").

By training the predicate classifier in SGG models with one-hot target labels, these neglected characteristics will respectively result in two adverse impacts: 1) It may lead to confused SGG predictions by simply assigning one-hot target labels without considering multiple reasonable predicates for each subject-object pair. Meanwhile, some over-confident predictions on the head categories also enlarge the performance gap between different categories. 2) It may lead to wrong predictions by directly assigning absolute \texttt{background} (\ie, one-hot target labels) to the missing-annotated subject-object pairs. Since the total probability is given to the wrong \texttt{background}, the model trusts the wrong label excessively.

To alleviate these issues, in this paper, we propose a novel model-agnostic Label Semantic Knowledge Distillation for unbiased SGG, dubbed \textbf{LS-KD}. Specifically, LS-KD integrates \emph{label semantic distributions} (LSD, \ie, the knowledge that reflects the correlations between subject-object instances and predicate categories) and original one-hot target labels to ``soft" labels, and uses these ``soft" labels to supervise the model training. Particularly, we utilize self-knowledge distillation (self-KD) to predict these LSDs, by transferring label semantic knowledge from a teacher model to the student model (\ie, the model used for final prediction). Meanwhile, we propose two different self-KD strategies: 1) \emph{Iterative self-KD}: the student model trained at an earlier step is utilized as the teacher model iteratively. 2) \emph{Synchronous self-KD}: the student and teacher models share the same backbone with two different classifiers, and both models are trained synchronously. Thanks to these designs, the student model can utilize the past and current predictions (\ie, LSD) of the teacher model in iterative and synchronous self-KD as informative supervision, respectively. Besides, we fuse the output of the teacher model with the original one-hot labels to generate soft target labels, which differs from other fusion strategies in the standard self-KD methods~\cite{furlanello2018born,zhang2020self,yuan2020revisiting}.

We evaluate our method on the most widely-used SGG benchmark: VG~\cite{krishna2017visual}. Since LS-KD is a model-agnostic debiasing method, it can be seamlessly incorporated into any advanced SGG architectures and consistently improve their performance. Extensive ablations and results on multiple SGG tasks and backbones have shown the effectiveness and generalization ability of LS-KD.

In summary, we make three main contributions in this paper:
\begin{enumerate}
    \item We are the first\footnote{For ``single-label classification", we mean that \emph{each triplet sample is assigned only one predicate label, and XE loss is used for training}. In contrast, all prior work only discussed that there may be more than one reasonable predicates for each subject-object pair.} to analyze the reason why the prevalent SGG training paradigm is unreasonable, \ie, the effect of treating predicate classification in SGG as a single-label classification.
    \item We propose a novel model-agnostic LS-KD, which can capture the correlations between the subject-object instances and different predicate categories.
    \item We conduct extensive experiments to prove the superiority of LS-KD, which can achieve a decent trade-off performance between different predicates.
\end{enumerate}

\begin{figure*}[!t]
  \centering
  \includegraphics[width=0.9\linewidth]{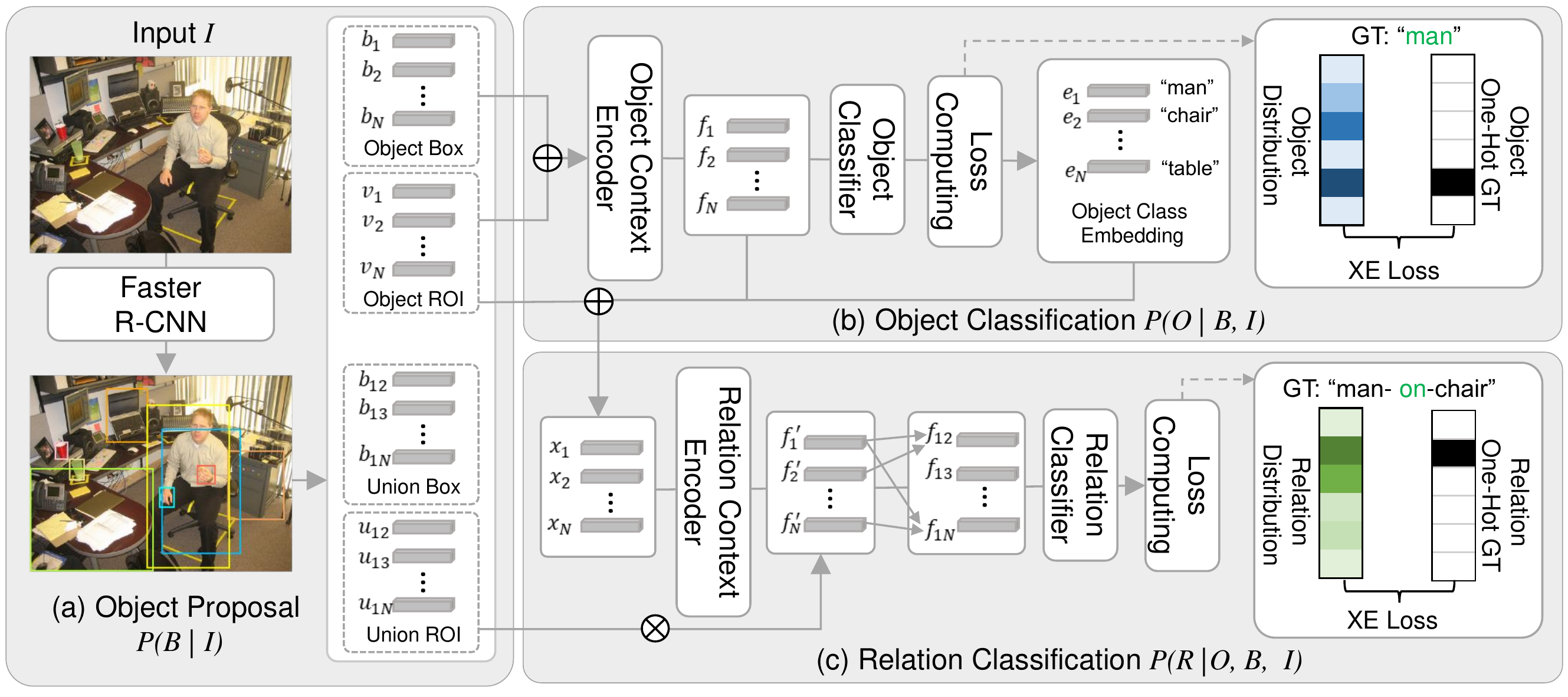}
  \caption{The framework of SGG baseline. a) Object proposal module generates the bounding box proposals of objects. b) Object classification module predicts the class probabilities of object proposals. c) Relation classification module predicts the relation class probabilities of pairwise objects. $\oplus$ denotes concatenation operation, $\otimes$ denotes element-wise multiplication. Dotted line only shows one instance for loss calculation.}
  \label{fig:framework_baseline}
\end{figure*}

\section{Related Work}

\noindent\textbf{Scene Graph Generation (SGG).}
Scene graphs are structural representations that transform raw visual input into semantic categories. After Lu \etal~\cite{lu2016visual}
firstly brought a large-scale dataset to attention, plentiful studies~\cite{tajrobehkar2021align,9726876} have been conducted in SGG. Existing SGG works can be divided into two groups: 1) Two-stage SGG: They first utilize a detector to obtain object proposals, then predict the classes of objects and their pairwise relations~\cite{li2017vip,xu2017scene,li2018factorizable,gu2019scene,zellers2018neural,yang2018graph,chen2019knowledge,tang2019learning,chen2019counterfactual}. In order to exploit the rich
global context, some works regard each image as a whole and adopt the fully connected graph~\cite{yang2018graph,chen2019knowledge}, the chained graph~\cite{zellers2018neural}, and the tree-structured graph~\cite{tang2019learning} to model the contexts among objects. 2) One-stage SGG: They use the fully convolutional network or Transformer to detect the objects and relations from image features directly~\cite{dong2021visual,liu2021fully}. In this paper, we propose a model-agnostic debiasing method that can be used in any SGG model. In our experiments, two popular two-stage SGG models (Motifs~\cite{zellers2018neural} \& VCTree~\cite{tang2019learning}) are used as baselines.

\noindent\textbf{Unbiased Scene Graph Generation.}
Due to the long-tailed distribution and other language bias issues~\cite{misra2016seeing} in SGG, numerous recent SGG work mainly focuses on unbiased SGG~\cite{tang2020unbiased,9829320}. They can be roughly divided into four categories: 1) Class-aware Re-sampling: They balance the data distribution by oversampling samples of tail categories or undersampling of head ones, or a combination of both~\cite{li2021bipartite}. Specially, Desai~\etal~\cite{desai2021learning} resample both predicate and entity categories to balance the distribution of triplets. 2) Loss Re-weighting: They assign different weights to different categories to reduce the contributions of the head categories in loss calculation. These designed weights are often based on prior knowledge~\cite{lin2017focal,knyazev2020graph} or learnable predicate correlations~\cite{yan2020pcpl}, and so on. 3) Biased-model-based: They try to separate unbiased predictions from biased trained models. For instance, Tang~\etal~\cite{tang2020unbiased} utilize counterfactual causality to separate biases from the biased model; Yu~\etal~\cite{yu2021cogtree} design a debiased reweight loss based on the biased model; Chiou~\etal~\cite{chiou2021recovering} recover the probability of unbiased prediction from the biased model. 4) Noisy Label Correction: They reformulate SGG as a noisy label learning problem and try to improve SGG annotation qualities. For example, Li~\etal~\cite{li2022devil,li2022nicest} propose the NICE and NICEST strategy that can not only detect noisy samples but also reassign more high-quality predicate labels. In this paper, we try to capture the possibility between the instance and multiple relation labels, which achieves a trade-off between the head and tail categories.

\noindent\textbf{Knowledge Distillation.}
Knowledge Distillation (KD) is a training strategy inspired by the ``teacher-student" learning behavior, in which the teacher model is the exporter of knowledge and the student model is the receiver~\cite{hinton2015distilling,9834142,gou2021knowledge,9257112,8890866}. KD that can effectively learn a lightweight student model from a cumbersome teacher model has been widely used in numerous scene understanding applications, \eg, VQA~\cite{chen2022rethinking}, scene graph generation~\cite{li2022integrating}. Traditional KD methods use knowledge from larger, better-performing teacher models to generate soft targets for student networks. Compared with conventional KD methods, self-knowledge distillation (self-KD) reduces the necessity of training a complicated teacher model in advance. It directly completes the distillation process with only the student network itself (\ie, the teacher model and student model share the same network), and it has shown conspicuous performance~\cite{yim2017gift,hahn2019self,yun2020regularizing,kim2021self}. Since self-KD can capture the correlations between categories~\cite{tang2020understanding} and it does not require any additional networks, in this paper, we design two self-KD strategies (\ie, iterative and synchronous) to predict the label semantic distribution.

\noindent\textbf{Label Smoothing and Label Confusion.} Label smoothing (LS) strategy~\cite{szegedy2016rethinking,zoph2018learning} is a regularization method, which has been widely used in classification tasks to prevent over-confident prediction and improve generalization ability. It typically combines a weighted uniform noise distribution with original one-hot target labels to obtain some softer labels for training. However, such soft labels generated merely by adding noise cannot reflect the correlations between each subject-object instance and multiple predicate categories. Label confusion (LC) method~\cite{guo2020label} is proposed in text classification task. It trains a label encoder to learn label representations and calculates the similarity between instances and labels to generate a better label distribution for training. Among these solutions, \cite{guo2020label} is the closest to our approach, but we do not need to design additional network encoders to capture correlations between instances and multiple predicates. In addition, some semi-supervised SGG methods also assign soft labels to unannotated samples~\cite{wang2020tackling,chen2019scene}.

\section{Approach}

\noindent\textbf{Scene Graph Definition.} Given an image $I$, a scene graph is formally
represented by $\mathcal{G}$ = \{ $\mathcal{N}$ = \{ $o_i$; $b_i$ \};  $\mathcal{E}$ = \{$r_{ij}$ \}\}, where the $\mathcal{N}$ and $\mathcal{E}$ denote the set of nodes and the set of edges, respectively. $o_i$ $\in$ $O$ denotes the object category of the $i$-th object, and its bounding box is $b_i$ $\in$ $B$. $r_{ij}$ $\in$ $R$ denotes the relation category between $i$-th object and $j$-th object. $B$, $O$, and $R$ represent the set of all predicted bounding boxes, object categories, and relation categories, respectively.

In this section, we first revisit the general SGG baseline in Section~\ref{sec:3.1}, then present details of the proposed Label Semantic Knowledge Distillation (LS-KD) in Section~\ref{sec:3.2}, and details of two self-KD strategies in Section~\ref{sec:3.3}.

\begin{figure*}[!t]
  \centering
  \includegraphics[width=0.85\linewidth]{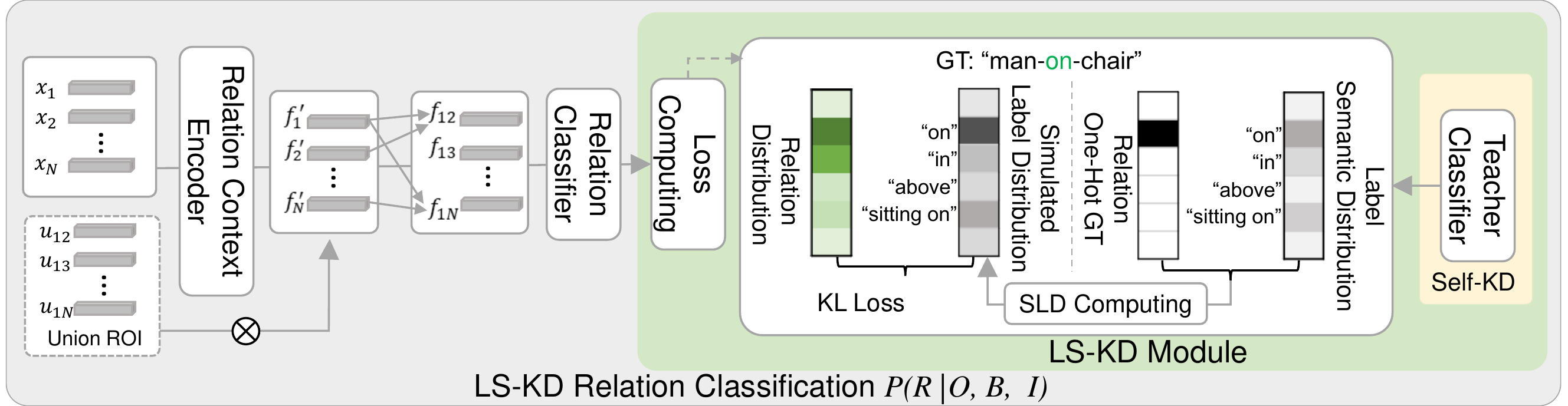}
  \caption{Overview of the LS-KD relation classification, which adds an LS-KD module to the baseline relation classification module. The green box represents the LS-KD module. The dotted line indicates an instance case when calculating loss.}
  \label{fig:lcm}
\end{figure*}

\subsection{Revisit Scene Graph Generation Baselines} \label{sec:3.1}

As shown in Fig.~\ref{fig:framework_baseline}, the most common SGG framework typically consists of three modules: object proposal, object classification and relation classification. These frameworks decompose the SGG target $P(\mathcal{G}|I)$ into three steps:
\begin{equation}
    P(\mathcal{G}|I) = P(B|I)P(O|B, I)P(R|O, B, I),
\end{equation}
where $P(B|I)$ is predicted by the object proposal module, $P(O|B, I)$ is predicted by the object classification module, and $P(R|B, O, I)$ is predicted by the relation classification module. In the following, we briefly introduce these modules.

\subsubsection{Object Proposal Module.} This module is to model $P(B|I)$. Given an image $I$, bounding box $b_i$ of each proposal is obtained by using a standard object detector (\eg, Faster R-CNN~\cite{ren2015faster}). Visual feature $v_i$ corresponding to each proposal region is extracted with RoIAlign operation~\cite{he2017mask}. Similarly, we compute the union box $b_{ij}$ of the $i$-th and $j$-th detected object region and extract the union visual features $u_{ij}$ in the same way. The union region provides contextual information about the interactions between objects to aid relation classification.

\subsubsection{Object Classification Module.} It is composed of an object context encoder and an object classifier to model $P(O|B,I)$. The object context encoder $\mathtt{Encoder_{obj}}$ takes the object ROI features \{$v_i$\} and bounding boxes \{$b_i$\} as input and outputs the object context representation \{$f_i$\}:
\begin{equation}
    f_i = \mathtt{{Encoder}_{obj}}(v_i \oplus b_i),
\end{equation}
where $\oplus$ denotes concatenation operation.
Some SGG work adopts bi-directional LSTMs~\cite{zellers2018neural},
bi-directional TreeLSTMs~\cite{tang2019learning} or fully connected layers~\cite{zhang2017visual,hung2020contextual} to encode context information among objects. The object classifier utilizes the fully-connected layer $\mathtt{FC}$ followed by a softmax operation with temperature $\tau$ to model $P(B|O,I)$ and predict object classes \{$o_i$\}:
\begin{gather}
    \mathtt{FC} (W,\sigma,f) = {W^T}f + \sigma,  \\
    \widehat{y^o_i} = \mathtt{Softmax}(\mathtt{FC}(W_o,f_i,\sigma_o),\tau),
\end{gather}
where $W$ and $\sigma$ are learnable weight and bias of the fully-connected layer, $\widehat{y^o_i}$ is the predicted label distribution of the object $o_{i}$.

\subsubsection{Relation Classification Module.} It consists of two parts: a relation context encoder and a relation classifier to model $P(R|O,B,I)$. The relation context encoder $\mathtt{{Encoder}_{rel}}$ takes the joint feature embeddings \{$x_i$\} as input and outputs the refined object features\{$f^{'}_i$\}:
\begin{gather}
     x_{i} = v_i \oplus f_i \oplus e_i, \\
     f^{'}_{i} = \mathtt{{Encoder}_{rel}}(x_{i}),
\end{gather}
where $e_i$ is the word embedding of object class $o_i$. The pairwise refined object features mixed with union feature $u_{ij}$ to obtain the relation representation $f_{ij}$ between the $i$-th and $j$-th objects. The relation classifier is similar to object classifier but with different categories to predict the label distribution:
\begin{gather}
    f_{ij} = [f^{'}_{i} \oplus f^{'}_{j}]\otimes u_{ij},\\
    \widehat{y^r_{ij}} = \mathtt{Softmax}(\mathtt{FC}(W_r,f_{ij},\sigma_r),\tau),
\end{gather}
where $\otimes$ denotes element-wise multiplication, $\widehat{y^r_{ij}}$ is the predicted label distribution of predicate $r_{ij}$.
$W_r$ and $\sigma_r$ are parameters of $\mathtt{FC}$ for relation classification.

\subsubsection{Training Strategy.} The object proposal module is frozen during training, while the two classification modules are trained using standard cross-entropy ($\mathtt{XE}$) loss $\mathcal{L}_{o}$ and $\mathcal{L}_{r}$ with one-hot object and relation target labels, respectively, \ie,
\begin{gather}
    \mathcal{L}_{o} = \mathtt{XE}(y^o, \widehat{y^o}) = -y^olog(\widehat{y^o}),  \\
    \mathcal{L}_{r} = \mathtt{XE}(y^r, \widehat{y^r}) = -y^rlog(\widehat{y^r}),
\end{gather}
where $y^o$ and $y^r$ denotes the object and relation one-hot target labels respectively, $\widehat{y^o}$ and $\widehat{y^r}$ is the predicted object and relation label distribution.

\begin{figure}[t]
  \centering
  \includegraphics[width=0.65\linewidth]{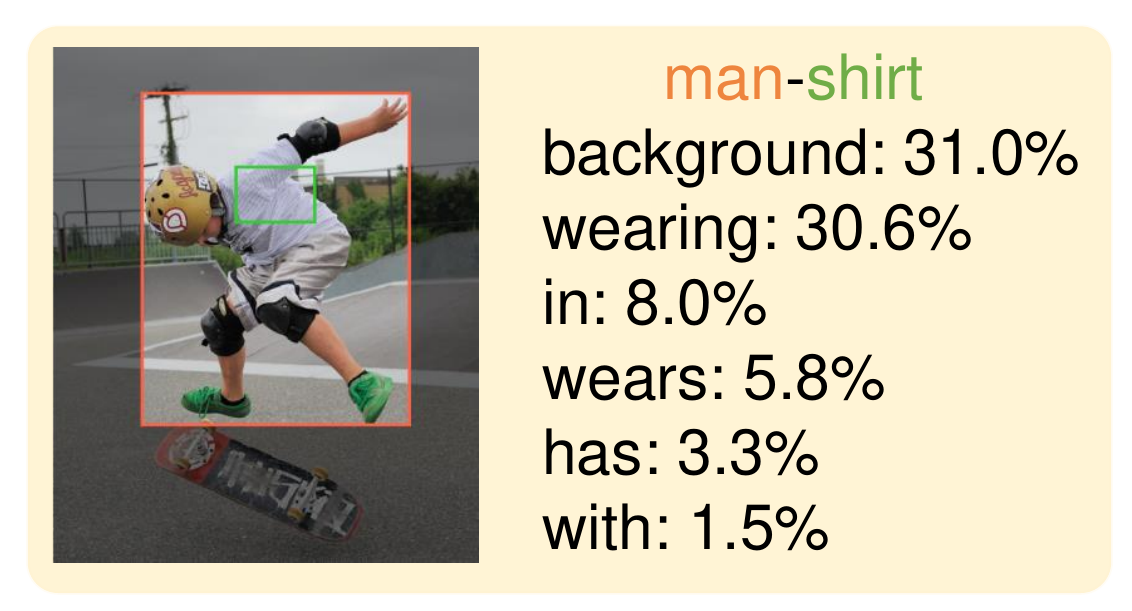}
  \caption{An example of the LSD of a missing-annotated triplet.}
  \label{bg}
\end{figure}

\begin{figure*}[t]
  \centering
  \includegraphics[width=0.8\linewidth]{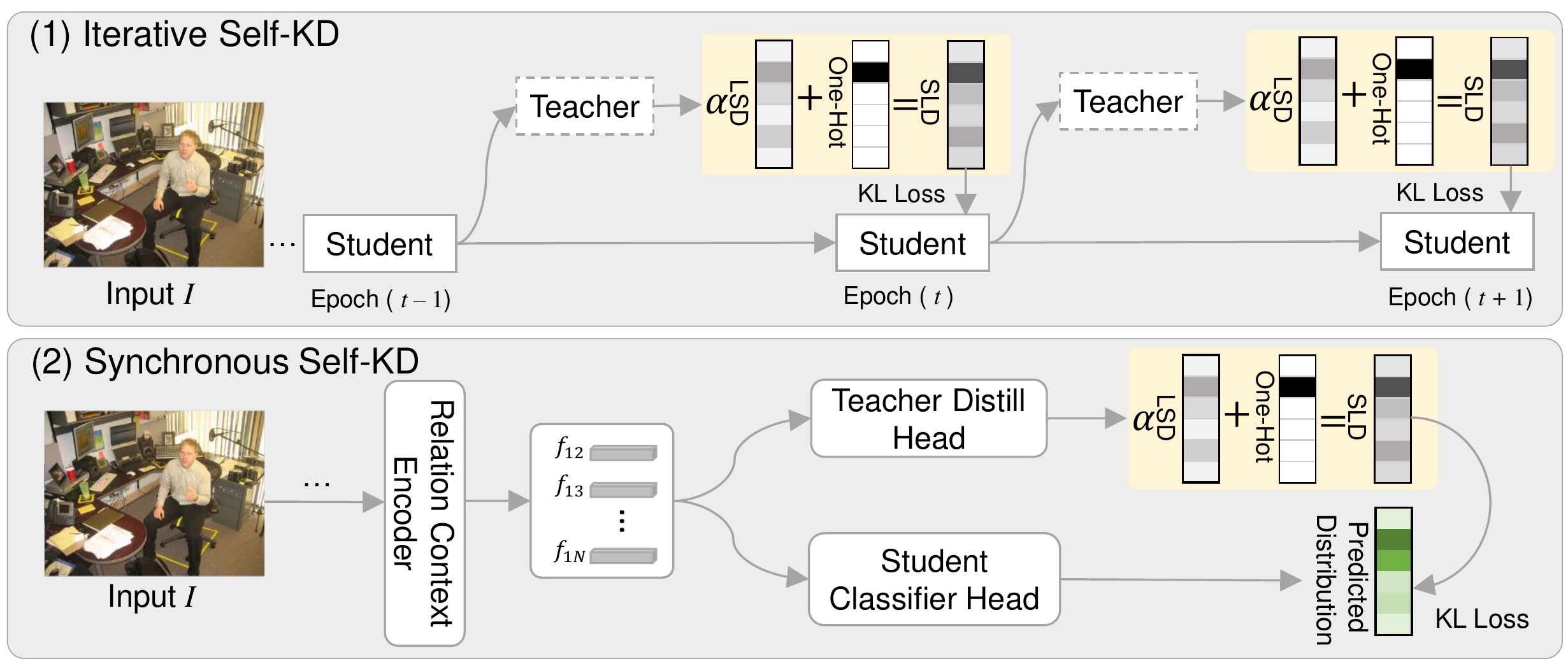}
  \caption{The pipeline of two self-Knowledge Distillation learning strategies. When training in $t$ epoch, student model trained at $t-1$ epoch becomes the teacher model. Dotted lines indicate that the model weights are frozen.}
  \label{fig:kd}
\end{figure*}

\subsection{LS-KD Relation Classification} \label{sec:3.2}

To alleviate the issues mentioned in Section~\ref{sec:1} caused by the training relation classifier just with one-hot target labels, we design an LS-KD relation classification with soft labels that consider the correlations between instances and multiple predicate categories. Taking triplet ``\texttt{man}-\texttt{on}-\texttt{chair}" in Fig.~\ref{fig:lcm} as an example, LS-KD can assign a portion of probability to predicate labels related to the ``\texttt{man}-\texttt{chair}", such as ``\texttt{sitting on}", ``\texttt{in}" and ``\texttt{above}" rather than give full probability to the one-hot label ``\texttt{on}". Compared to the basic relation classification, the LS-KD relation classification adds an extra LS-KD module.

The label semantic knowledge distillation module is constructed by an output of the teacher model classifier trained by self-KD and a simulated label distribution computing block. The details of self-KD learning will be described in Section~\ref{sec:3.3}. The teacher model classifier takes the relation representations \{$f_{ij}$\} extracted by the relation encoder as an input and outputs Label Semantic Distributions (LSDs). The LSD can dynamically capture the correlations among multiple predicate categories for each instance. Finally, we adopt a controlling hyperparameter $\alpha$ to merge the original one-hot vector and LSD and then utilize a softmax function to generate the final Simulated Label Distribution (SLD). The SLD is used as the soft target label to replace the one-hot vector to supervise the training of the relation classifier. The hyperparameter $\alpha$ determines the proportion of the final fusion between the LSDs and the original one-hot target labels:
\begin{gather}
    y^{t}_{ij} =  \mathtt{Softmax}(\mathtt{FC}(W,f_{ij},\sigma),\tau),\\
    y^{s}_{ij} = \mathtt{Softmax}(\alpha y^{r}_{ij} + y^{t}_{ij}, \tau),
\end{gather}
where $y^{t}_{ij}$ is the output LSD of the teacher model classifier, $y^{r}_{ij}$ is the original one-hot vector of the predicate label, and $y^{s}_{ij}$ is the computed SLD. With the early supervision of the original single label, the output of the teacher model can reflect the correlations between each subject-object pair and multiple predicate categories. However, the probability gap between head and tail categories is large due to the long-tail distribution, which results in excessive suppression of reasonable tail predicates. Thus, we let the model learn more reasonable $y^{t}_{ij}$ that can soften the original labels to reduce the gap between head and tail predicates. Meanwhile, the supervision in our method is the pseudo label distribution $y^{s}_{ij}$.

During training, we adopt KL-divergence~\cite{kullback1951information} loss to measure the similarity between two probability distributions (the SLD $y^s_{ij}$ and the predicted relation label distribution $\widehat{y^r_{ij}}$):

\begin{equation}
\mathcal{L}_r = \mathtt{KL}(y^s,\widehat{y^r})=\sum\limits_c^C {y_c^s\log (\frac{{y^s_c}}{{\widehat{y^r_c}}})},
\end{equation}
where $c \in C$ represents $c$-th predicate category. Through the training of the LS-KD module, the soft target dynamically changes for each triplet instance based on the semantic knowledge learned by the teacher model among predicate labels.

The learned simulated label distribution not only contributes to better representing the instance with multiple confusing predicate labels, but also it is more robust to missing-annotated triplets. For the missing-annotated subject-predicate-object triplets, a portion of the probability of the original one-hot \texttt{background} label is allocated to some reasonable foreground predicate labels (\eg, ``\texttt{wearing}", ``\texttt{in}", and other predicates for the ``\texttt{man}-\texttt{chair}" pair in Fig.~\ref{bg}), thus the model can still learn some label semantic knowledge.

\subsection{Self-Knowledge Distillation for LSD} \label{sec:3.3}

Since the probabilistic output of the teacher model in self-KD can reflect the correlations of multiple predicate labels for each instance, it does not require the design of an extra model structure (\ie, the student and teacher models share the same network).
In this paper, we explore two different self-KD strategies: iterative self-KD and synchronous self-KD, to prompt the teacher model to predict label semantic distributions. The pipelines are shown in Fig.~\ref{fig:kd}.

\subsubsection{Iterative Self-Knowledge Distillation.} During the training, the student model becomes the teacher model itself iteratively. By iterative self-KD, the student model can utilize its past predictions to serve more informative supervisions (\ie, label semantic knowledge) for training. Concretely, the student model of the ($t$-1)-th epoch becomes the teacher model during training in ($t$)-th epoch, and the parameters of the teacher model are frozen. The outputs of the teacher model are fused with one-hot target labels by hyperparameter $\alpha$ in the LS-KD module to generate soft labels. These soft labels are utilized to supervise the student model training with KL loss. Especially when training at the first epoch, we just utilize a cross-entropy loss with the one-hot label to train the student model.

\subsubsection{Synchronous Self-Knowledge Distillation.} During the training, the student model and the teacher model share the same relation encoder. However, they adopt two pseudo-siamese classifier heads (\ie, the networks with the same structure but different parameters): the teacher distill head and the student classifier head. These two heads take the same relation representations \{$f_{ij}$\} as input. The difference is that the teacher distill head is trained by cross-entropy loss with one-hot labels. Similarly, the output of the teacher distill head is used in the LS-KD to generate soft target labels to supervise the training of the student classifier head by KL loss.

\noindent\textbf{Comparison with Prior Self-KD Models.} Our fusion strategy of the original one-hot label and the output of teacher model differs from all the standard self-KD methods~\cite{furlanello2018born,zhang2020self,yuan2020revisiting}. More importantly, our motivation and main contributions are to model the correlation between multiple predicates for each sub-obj instance. To achieve the goal, we resorted to KD to generate soft labels for supervised training. In addition, to avoid designing additional model structures, we adopted self-KD strategy and proposed two variants.

\section{Experiments}

\subsection{Experimental Settings and Details}

\noindent\textbf{Datasets.} We conducted the experiments on the most prevalent SGG benchmark: Visual
Genome (VG)~\cite{krishna2017visual} that totally contains 108,073 images. Since the number of samples in most predicate categories is quite limited, in our experiments, we followed the widely-used data splits~\cite{xu2017scene}, which consists of 150 object categories and 50 predicate categories. The dataset is split with 70\% of images as the training set and the other 30\% as the testing set, respectively. Following the same protocol in~\cite{zellers2018neural}, We sampled 5000 images from the training set as the validation set for hyperparameter tuning.

\noindent\textbf{Tasks.} We evaluated the model on the three SGG tasks~\cite{xu2017scene}: 1) \emph{Predicate Classification} (\textbf{PredCls}): Given the ground-truth object bounding boxes with ground-truth object classes, the model is required to predict the predicate
categories of object pairs. 2) \emph{Scene Graph Classification} (\textbf{SGCls}): Given the ground-truth object bounding boxes without classes, the model is required to predict both the object categories and predicate categories. 3) \emph{Scene Graph Generation} (\textbf{SGGen}): Given an image, the model is required to detect all object bounding boxes, and predict categories of both the object and predicate.

\noindent\textbf{Metrics.} We evaluated all results on three metrics: 1) \emph{Recall@K} (\textbf{R@K}): It calculates the fraction of the top $K$ confident predicted subject-relation-object triplets that match the ground-truth. In this paper, we utilized $K = \{50, 100\}$. 2) \emph{mean Recall@K} (\textbf{mR@K}): It retrieves each predicate separately and then averages the R@K for all predicates, we also reported $K = \{50, 100\}$. 3) \textbf{Mean}: It is the average of all mR@K and R@K scores. Since R@K favors head predicates and mR@K favors tail predicates, the Mean metric is better for evaluating the performance across all predicates. For unbiased SGG, mR@K is often boosted at the cost of the drop in R@K (\ie, many state-of-the-art debiasing models~\cite{chiou2021recovering,tang2020unbiased,yu2021cogtree} suffer from obvious performance drops on head predicates), and thus Mean is a comprehensive metric to measure the performance on different predicate categories~\cite{li2022devil}.

\noindent\textbf{Implementation Details.} Following the similar protocol in~\cite{tang2020unbiased}, we adopted Faster R-CNN with the ResNeXt-101-FPN~\cite{lin2017feature} backbone to detect object bounding boxes and extract RoI features, and kept the parameters frozen during the training. We utilized the detector trained by~\cite{tang2020unbiased}, which achieved 28.14 mAP on the VG test set (\ie, using 0.5 IoU threshold). The hyperparameter $\alpha$ utilized in LS-KD relation classification was set to 4, and temperature $\tau$ of two self-KD were set to 1. The confidence of a predicted triplet is calculated by multiplying the classification score of three components (\ie, subject, object and predicate). The models in this paper were trained by SGD optimizer, where the mini-batch size was 12, and the initial learning rate was 0.01. After the validation performance plateaus, the learning rate would be decayed by 10 two times. All the experiments were implemented with PyTorch and conducted with NVIDIA 2080ti GPUs.

\subsection{Comparison with State-of-the-Arts}
\begin{table*}[!t]
    \centering
    \caption{Performance (\%) of state-of-the-art SGG models on three SGG tasks. ``Mean" is the average of mR@50/100 and R@50/100.}
        \scalebox{1}{
            \begin{tabular}{l |  c c  |c | c  c  |c |  c c  |c}
                \hline
                \multirow{2}{*}{Models} & \multicolumn{3}{c|}{PredCls} & \multicolumn{3}{c|}{SGCls} & \multicolumn{3}{c}{SGGen} \\
                \cline{2-10}
                & \scriptsize{mR@50/100\ } & \scriptsize{\ R@50/100\ } & \scriptsize{Mean }
              & \scriptsize{mR@50/100\ } & \scriptsize{\ R@50/100\ } & \scriptsize{Mean }
              & \scriptsize{mR@50/100\ } & \scriptsize{\ R@50/100\ } & \scriptsize{Mean } \\
              \hline
                MSDN~\cite{li2017scene}
                & 15.9/17.5 & 64.6/66.6 & 41.2
                & 9.3/9.7  & 38.4/39.8 & 24.3
                & 6.1/7.2  & 31.9/36.6 & 20.5 \\
                G-RCNN~\cite{yang2018graph}
                & 16.4/17.2 & 64.8/66.7 & 41.3
                & 9.0/9.5  & 38.5/37.0 & 23.5
                & 5.8/6.6  & 29.7/32.8 & 18.7 \\
                DT2ACBS~\cite{desai2021learning}
                & 35.9/39.7 & 23.3/25.6 & 31.1
                & 24.8/27.5 & 16.2/17.6 & 21.5
                & 22.0/24.4 & 15.0/16.3 & 19.4 \\

                \hline
                \footnotesize{Motifs+TDE~\cite{tang2020unbiased}}
                & 24.2/27.9 & 45.0/50.6 & 36.9
                & 13.1/14.9 & 27.1/29.5 & 21.2
                & 9.2/11.1 & 17.3/20.8 & 14.6 \\
                \footnotesize{Motifs+PCPL~\cite{yan2020pcpl}}
                & 24.3/26.1 & 54.7/56.5 & 40.4
                & 12.0/12.7 &  35.3/36.1 & 24.0
                & 10.7/12.6 & 27.8/31.7 & \textbf{20.7} \\
                \footnotesize{Motifs+CogTree~\cite{yu2021cogtree}}
                & 26.4/29.0 & 35.6/36.8	& 32.0
                & 14.9/16.1 & 21.6/22.2 & 18.7 	
                & 10.4/11.8 & 20.0/22.1 & 16.1 \\
                \footnotesize{Motifs+DLFE~\cite{chiou2021recovering}}
                & 26.9/28.8 & 52.5/54.2 & 40.6
                & 15.2/15.9 & 32.3/33.1 & 24.1
                & 11.7/13.8 & 25.4/29.4 & 20.1 \\
                \footnotesize{Motifs+BPL-SA~\cite{guo2021general}}
                &  29.7/31.7 & 50.7/52.5 & 41.2
                &  16.5/17.5 & 30.1/31.0 & 23.8
                &  13.5/15.6 & 23.0/26.9 & 19.8 \\
                \footnotesize{\cellcolor{mygray-bg}{\textbf{Motifs+LS-KD(Iter)}}}
                & \cellcolor{mygray-bg}{24.1}/\cellcolor{mygray-bg}{27.4} &  \cellcolor{mygray-bg}{55.1}/\cellcolor{mygray-bg}{59.1} & \cellcolor{mygray-bg}\textbf{41.4}& \cellcolor{mygray-bg}{13.8}/\cellcolor{mygray-bg}{15.2}& \cellcolor{mygray-bg}{33.8}/\cellcolor{mygray-bg}{35.8} &
                \cellcolor{mygray-bg}{24.7}	& \cellcolor{mygray-bg}{9.7}/\cellcolor{mygray-bg}{11.5}	&  \cellcolor{mygray-bg}{25.6}/\cellcolor{mygray-bg}{29.7} & \cellcolor{mygray-bg}{19.1} \\
                \footnotesize{\cellcolor{mygray-bg}{\textbf{Motifs+LS-KD(Syn)}}}
                & \cellcolor{mygray-bg}{22.3}/\cellcolor{mygray-bg}{25.3} &  \cellcolor{mygray-bg}{56.9}/\cellcolor{mygray-bg}{60.3} & \cellcolor{mygray-bg}{41.2}& \cellcolor{mygray-bg}{12.4}/\cellcolor{mygray-bg}{14.1}& \cellcolor{mygray-bg}{35.4}/\cellcolor{mygray-bg}{37.1} &
                \cellcolor{mygray-bg}{\textbf{24.8}}	& \cellcolor{mygray-bg}{8.1}/\cellcolor{mygray-bg}{10.0}	&  \cellcolor{mygray-bg}{27.5}/\cellcolor{mygray-bg}{31.6} & \cellcolor{mygray-bg}{19.3} \\
                \hline

                \footnotesize{VCTree+TDE~\cite{tang2020unbiased}}
                & 26.2/29.6 & 44.8/49.2 & 37.5
                & 15.2/17.5 & 28.8/32.0 & 23.4
                & 9.5/11.4 & 17.3/20.9 & 14.8 \\
                \footnotesize{VCTree+PCPL~\cite{yan2020pcpl}}
                & 22.8/24.5 &  56.9/58.7 & 40.7
                & 15.2/16.1 &  40.6/41.7 & 28.4
                & 10.8/12.6 & 26.6/30.3 & \textbf{20.1} \\
                \footnotesize{VCTree+CogTree~\cite{yu2021cogtree}}
                & 27.6/29.7 & 44.0/45.4 & 36.7
                & 18.8/19.9 & 30.9/31.7 & 25.3
                & 10.4/12.1 & 18.2/20.4 & 15.3 \\
                \footnotesize{VCTree+DLFE~\cite{chiou2021recovering}}
                & 25.3/27.1 & 51.8/53.5 & 39.4
                & 18.9/20.0 & 33.5/34.6 & 26.8
                & 11.8/13.8 & 22.7/26.3 & 18.7 \\
                \footnotesize{VCTree+BPL-SA~\cite{guo2021general}}
                &  30.6/32.6 & 50.0/51.8 & 41.3
                &  20.1/21.2 & 34.0/35.0 & 27.6
                &  13.5/15.7 & 21.7/25.5 & 19.1 \\
                \footnotesize{\cellcolor{mygray-bg}{\textbf{VCTree+LS-KD(Iter)}}}
                & \cellcolor{mygray-bg}{24.2}/\cellcolor{mygray-bg}{27.1} &  \cellcolor{mygray-bg}{55.7}/\cellcolor{mygray-bg}{59.5} & \cellcolor{mygray-bg}{\textbf{41.6}}& \cellcolor{mygray-bg}{17.3}/\cellcolor{mygray-bg}{19.1}& \cellcolor{mygray-bg}{38.4}/\cellcolor{mygray-bg}{40.5} &
                \cellcolor{mygray-bg}\textbf{28.8}	& \cellcolor{mygray-bg}{9.7}/\cellcolor{mygray-bg}{11.3}	&  \cellcolor{mygray-bg}{25.3}/\cellcolor{mygray-bg}{29.5} & \cellcolor{mygray-bg}{19.0} \\
                \footnotesize{\cellcolor{mygray-bg}{\textbf{VCTree+LS-KD(Syn)}}}
                & \cellcolor{mygray-bg}{22.0}/\cellcolor{mygray-bg}{25.4} &  \cellcolor{mygray-bg}{57.5}/\cellcolor{mygray-bg}{60.8} & \cellcolor{mygray-bg}{41.4}& \cellcolor{mygray-bg}{14.6}/\cellcolor{mygray-bg}{16.5}& \cellcolor{mygray-bg}{40.6}/\cellcolor{mygray-bg}{42.6} &
                \cellcolor{mygray-bg}{28.6}	& \cellcolor{mygray-bg}{8.2}/\cellcolor{mygray-bg}{10.1}	&  \cellcolor{mygray-bg}{27.5}/\cellcolor{mygray-bg}{31.2} & \cellcolor{mygray-bg}{19.3} \\
                \hline
            \end{tabular}
        }
    \label{tab:compare_with_sota}
\end{table*}

\begin{figure*}[!t]
    \centering
    \includegraphics[width=0.8\linewidth]{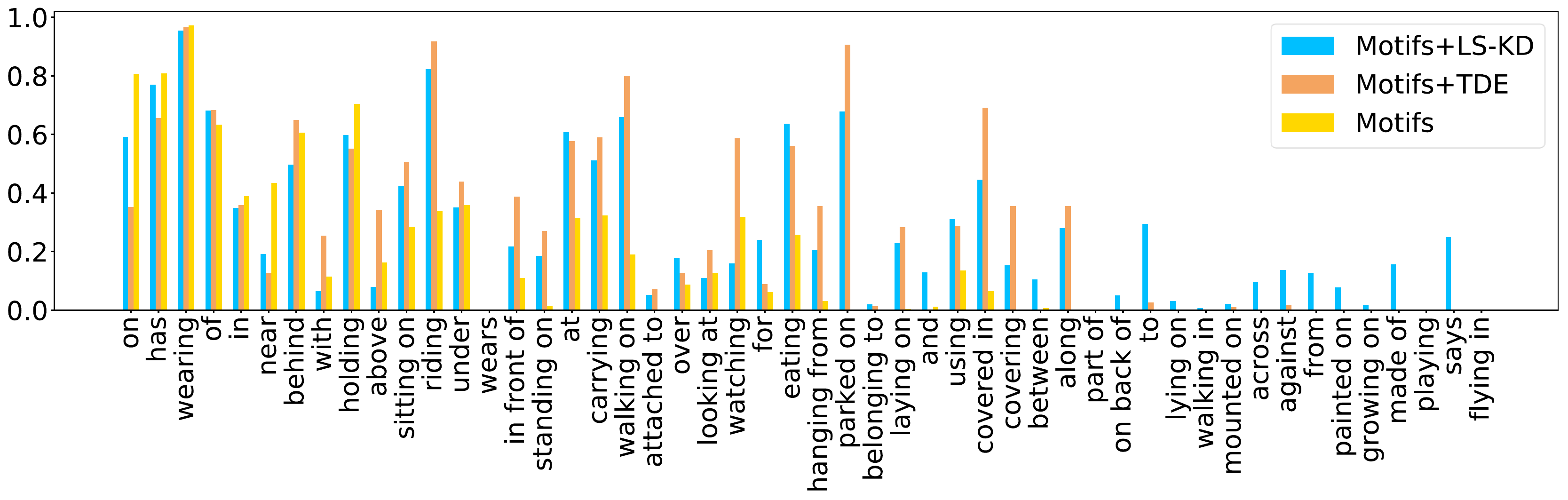}
    \vspace{-1em}
    \caption{Recall statistics for each predicate category. The baseline model is Motifs~\cite{zellers2018neural}.}
    \label{fig:stat}
\end{figure*}

\textbf{Settings.} We conducted LS-KD with iterative and synchronous self-KD on the two popular baselines (Motifs~\cite{zellers2018neural} and VCTree~\cite{tang2019learning}), denoted as \textbf{LS-KD (Iter)} and \textbf{LS-KD (Syn)} respectively. In Table~\ref{tab:compare_with_sota}, we listed the results of the various state-of-the-art SGG models for reference, and divided them into two groups: 1) The methods are not model-agnostic debiasing models, including MSDN~\cite{li2017scene}, G-RCNN~\cite{yang2018graph}, and DT2ACBS~\cite{desai2021learning}.
2) The methods are model-agnostic debiasing strategies, including TDE~\cite{tang2020unbiased}, PCPL~\cite{yan2020pcpl}, CogTree~\cite{yu2021cogtree}, DLFE~\cite{chiou2021recovering}, and BP-LSA~\cite{guo2021general}. Since LS-KD is model-agnostic, we mainly compared with methods in the second group with the same baselines. To further illustrate the performance of our model, we show recall statistics for each predicate of LS-KD, TDE~\cite{tang2020unbiased}, and Motifs~\cite{zellers2018neural} in Fig.~\ref{fig:stat} for comparisons. Besides, we report the R@100 metric of each predicate category group under the PredCls setting in Table~\ref{table:group}.

\begin{table}[!t]
    \centering
    \caption{Recall@100 of each group,\ie, head, body and tail, under PredCls setting. MEAN is the average of three groups.}
    \begin{tabular}{|l|c|c|c|c|}
    \hline
    \multirow{2}{*}{Models} & \multicolumn{4}{c|}{PredCls R@100} \\
    \cline{2-5}
    & \small{Head}  & \small{Body}  & \small{Tail}  & \small{MEAN} \\
    \hline
    Motifs~\cite{zellers2018neural} & 80.5  & 29.3  & 5.7   & 42.3 \\
    ~~+TDE~\cite{tang2020unbiased} & 66.4  & 40.1  & 20.3  & 38.5 \\
    ~~+\textbf{LS-KD} & 74.9  & 35.5  & 20.5  & \textbf{43.6} \\
    \hline
    \end{tabular}
    \label{table:group}
\end{table}

\noindent\textbf{Results.}
From the results in Table~\ref{tab:compare_with_sota}, we find that: 1) LS-KD exceeds the performance of baseline models under all the three settings and achieves comparable performance with some SOTA methods (\eg, TDE, PCPL and DLFE) in the mR@K metrics. 2) LS-KD achieves a better trade-off between the R@K and mR@K, and surpasses the state-of-the-art method in the Mean metric under both PredCls and SGCls settings (\eg, 41.2\% in BPL-SA \vs~41.4\% in LS-KD, 23.8 \% in BPL-SA \vs~24.8\% in LS-KD based on Motifs). To the best of our knowledge, all SOTA debiasing SGG models have lower performance than the baseline for the Mean metric. In contrast, our LS-KD has the least performance drops on the head predicates (\cf~R@K) and keeps the performance of tail predicates (\cf~mR@K), which proves the superiority of LS-KD by synthesizing all predicates. 3) LS-KD (Syn) can achieve higher performance than LS-KD (Iter) on R@K, because the teacher head of LS-KD (Syn) is trained by cross-entropy loss with one-hot target labels, which is susceptible to the influence of long-tailed distributions. Furthermore, as shown in Fig.~\ref{fig:stat} and Table~\ref{table:group}, we can observe that our proposed LS-KD could achieve a decent trade-off among different predicate categories (\eg, 38.5\% in TDE \vs~43.6\% in LS-KD in the Mean metric under PredCls settings based on Motifs).

\subsection{Comparison with Label Smoothing and Confusion}

\begin{table}[!t]
    \centering
    \caption{Comparison with LS and LC. The baseline model is Motifs~\cite{zellers2018neural}.}
    \scalebox{0.98}{
        \begin{tabular}{|l|c|c|c|}
        \hline
         & \multicolumn{3}{c|}{PredCls} \\
        \cline{2-4}
        Methods & \small{mR@50/100} & \small{R@50/100}  & \small{Mean} \\
        \hline
        Motifs+Label Smooth (LS) & 24.6 / 27.9 & 50.6 / 55.8 & 39.7 \\
        Motifs+Label Confusion (LC) & 25.1 / 28.6 & 50.3 / 55.3 & 39.8 \\
        \textbf{Motifs+LS-KD} & 24.1 / 27.4 & 55.1 / 59.1 &	\textbf{41.4} \\
        \hline
        \end{tabular}%
    }
    \label{tab:lslm}%
\end{table}

\noindent\textbf{Settings.} In Table~\ref{tab:lslm}, we re-implemented the general Label Smoothing (LS)~\cite{wang2021proselflc} and Label Confusion (LC)~\cite{guo2020label} methods under the PredCls setting for SGG. LS softens one-hot targets by adding uniform distribution. LC captures semantic overlap among labels by calculating the similarity between instances and labels, and generates soft targets for training. For a fair comparison, all the $\alpha$ controlling the synthesis of the original one-hot vector and soft distribution were set to 4.

\noindent\textbf{Results.} From the results, we get the following findings: 1) By making the model less confident during training, the LS method can alleviate the over-fitting of the model to a certain extent, especially for the head predicate classes that account for a large proportion. LS only softens the targets by adding uniformly distributed noise. Thus, it cannot reflect the correlations among predicates labels, resulting in poor performance on Mean (39.7\% in LS \vs~41.4\% in LS-KD). 2) The LC method can achieve better results on mR@K, but it is inferior to our method in the performance on Mean (39.8\% in LC \vs~41.4\% in LS-KD) that considers all predicates synthetically. Besides, it needs to design additional networks to encode the representation of labels, and it takes a lot of time to calculate the similarity among all labels for each instance. 3) Both LS and LC methods can significantly improve the performance in the mR@K compared with the baseline performance, which means LS and LC are helpful even without using a teacher-student model. However, since our motivation is to generate soft labels that can reflect the correlations between subject-object pairs and multiple predicates, we adopt KD whose output is a probability distribution. Meanwhile, as the training target is soft labels, it cannot depart from multi-label paradigm.

\subsection{Ablation Studies}
\subsubsection{Ablation Study on $\alpha$ in LS-KD.} In Table~\ref{tab:alpha}, we set various hyperparameters $\alpha$ in LS-KD to control the degree of fusion between the LSD learned by the teacher model and the original one-hot vector. The larger $\alpha$, the larger weight will be assigned to the one-hot vector when generating a soft label in the SLD module.

When the hyperparameter $\alpha$ is set to 5, the original one-hot vector takes a larger proportion than the other two values (\ie, 3 and 4), and the model makes over-confident predictions on head predicate with larger sample size, so the R@K is relatively high. When the hyperparameter $\alpha$ is set to 3, the LSD accounts for a greater proportion than the other two values (\ie, 4 and 5). If the instance is related to multiple labels, the sample will not excessively fit just one label, and the probability gap between head and tail predicate categories will decrease relatively. Therefore, mR@K is relatively high.
\begin{table}[!t]
    \centering
    \makeatletter\def\@captype{table}\makeatother\caption{Ablation studies on the hyperparameter $\alpha$ of LS-KD with iterative self-KD and synchronous self-KD. The baseline model is Motifs~\cite{zellers2018neural}.}
    \scalebox{1.0}{
        \begin{tabular}{|c|c|c|c|c|}
        \hline
         & & \multicolumn{3}{c|}{PredCls} \\
        \cline{3-5}
        \ $\alpha$\ & Type & \small{mR@50/100} & \small{R@50/100}  & \small{Mean} \\
        \hline
        - & - &16.5 / 17.8 & 65.5 / 67.2 & 41.8 \\
        \hline
        \multirow{2}{*}{3}
        & Iter & 25.6 / 29.6 & 43.5 / 49.8 & 37.1 \\
        & Syn & 24.2 / 28.3 & 44.9 / 51.0 & 37.1 \\
        \hline
        \multirow{2}{*}{4}
        & Iter & 24.1 / 27.4 & 55.1 / 59.1 & 41.4 \\
        & Syn & 22.3 / 25.3 & 56.9 / 60.3 & 41.2 \\
        \hline
        \multirow{2}{*}{5}
        & Iter & 20.6 / 23.0 & 60.7 / 63.5 & 42.0 \\
        & Syn & 18.8 / 21.0 & 61.7 / 64.1 & 41.4 \\
        \hline
        \end{tabular}%
    }
    \label{tab:alpha}%
\end{table}

\begin{table}[!t]
    \centering
    \makeatletter\def\@captype{table}\makeatother\caption{Ablation studies on the temperature $\tau$ of LS-KD with iterative self-KD and synchronous self-KD. The baseline model is Motifs~\cite{zellers2018neural}.}
    \scalebox{1.0}{
        \begin{tabular}{|c|c|c|c|c|}
        \hline
         & & \multicolumn{3}{c|}{PredCls} \\
        \cline{3-5}
        \ $\tau$\ & Type & \small{mR@50/100} & \small{R@50/100}  & \small{Mean} \\
        \hline
        - & - &16.5 / 17.8 & 65.5 / 67.2 & 41.8 \\
        \hline
        \multirow{2}{*}{0.1}
        & Iter & 23.6 / 26.7 & 55.7 / 59.7 & 41.4 \\
        & Syn & 21.1 / 24.1 & 57.1 / 60.6 & 40.7 \\
        \hline
        \multirow{2}{*}{1}
        & Iter & 24.1 / 27.4 & 55.1 / 59.1 & 41.4 \\
        & Syn & 22.3 / 25.3 & 56.9 / 60.3 & 41.2 \\
        \hline
        \multirow{2}{*}{10}
        & Iter & 24.2 / 28.0 & 51.7 / 57.3 & 40.3 \\
        & Syn & 22.3 / 26.0 & 51.9 / 56.9 & 39.3 \\
        \hline
        \end{tabular}%
    }
    \label{tab:t}%
\end{table}

\begin{figure}[!t]
    \centering
    \includegraphics[width=\linewidth]{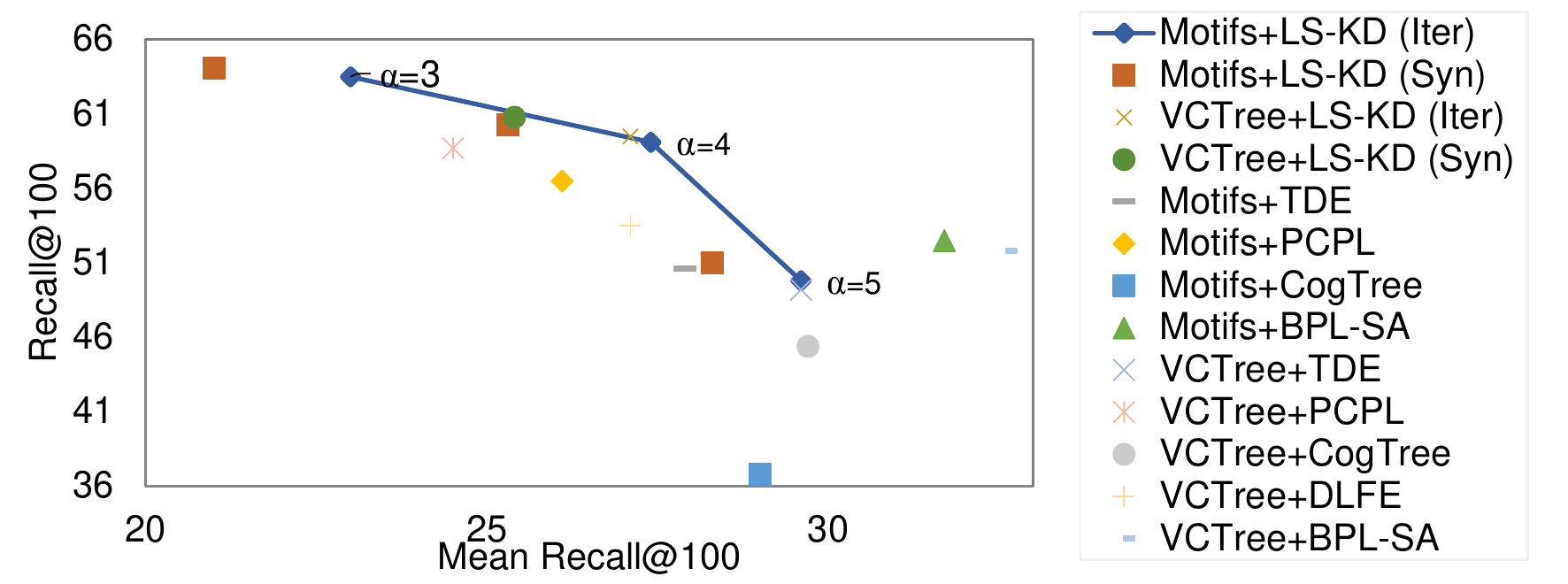}
    \caption{The mR-R curve of SOTA models under PredCls setting.}
    \label{fig:mR_R}
\end{figure}

Also, since $\alpha$ is a hyperparameter for the trade-off between Recall and Mean Recall performance, we plot the mR-R curve of all the operating results in Fig.~\ref{fig:mR_R}. As shown in Fig.~\ref{fig:mR_R}, almost all prior debiasing methods are under this curve, which confirms the effectiveness of our proposed method.
\subsubsection{Ablation Study on the Temperature $\tau$ in LS-KD.} In Table~\ref{tab:t}, we set the different temperature $\tau$ of softmax function in LS-KD. The larger $\tau$ is, the smoother the output probability distribution of softmax is. From the results, we find that the smoother the LSD, the better the performance on mR@K, and the steeper the LSD, the better the performance on R@K.

\begin{table}[!t]
    \centering
    \caption{Ablation studies on the iterative interval $t$ of iterative self-KD. The baseline model is Motifs~\cite{zellers2018neural}. 0.5 iteration means the iteration period is 0.5 epoch.}
        \scalebox{1.0}{
            \begin{tabular}{|c|c|c|c|}
            \hline
             & \multicolumn{3}{c|}{PredCls} \\
            \cline{2-4}
            $t$ & \small{mR@50/100} & \small{R@50/100}  & \small{Mean} \\
            \hline
            - & 16.5 / 17.8 & 65.5 / 67.2 & 41.8 \\
            0.5 & 24.0 / 27.3 & 54.6 / 58.9 & 41.2 \\
            1 & 24.0 / 27.2 & 54.8 / 59.2 &	41.3 \\
            2 & 24.1 / 27.4 & 55.1 / 59.1 & 41.4 \\
            \hline
            \end{tabular}%
        }
        \label{tab:iterval}%
\end{table}

\subsubsection{Ablation Study on Iterative Interval $t$ of Iterative Self-KD.} We conducted experiments with different iterative intervals $t$ of iterative self-KD (\ie, the interval of iterations to transform the student model into the teacher model). Observing the results, we can find that the larger the iteration interval, the higher the performance on Mean.

\subsection{Qualitative Results}

\begin{figure*}[!t]
  \centering
  \includegraphics[width=0.9\linewidth]{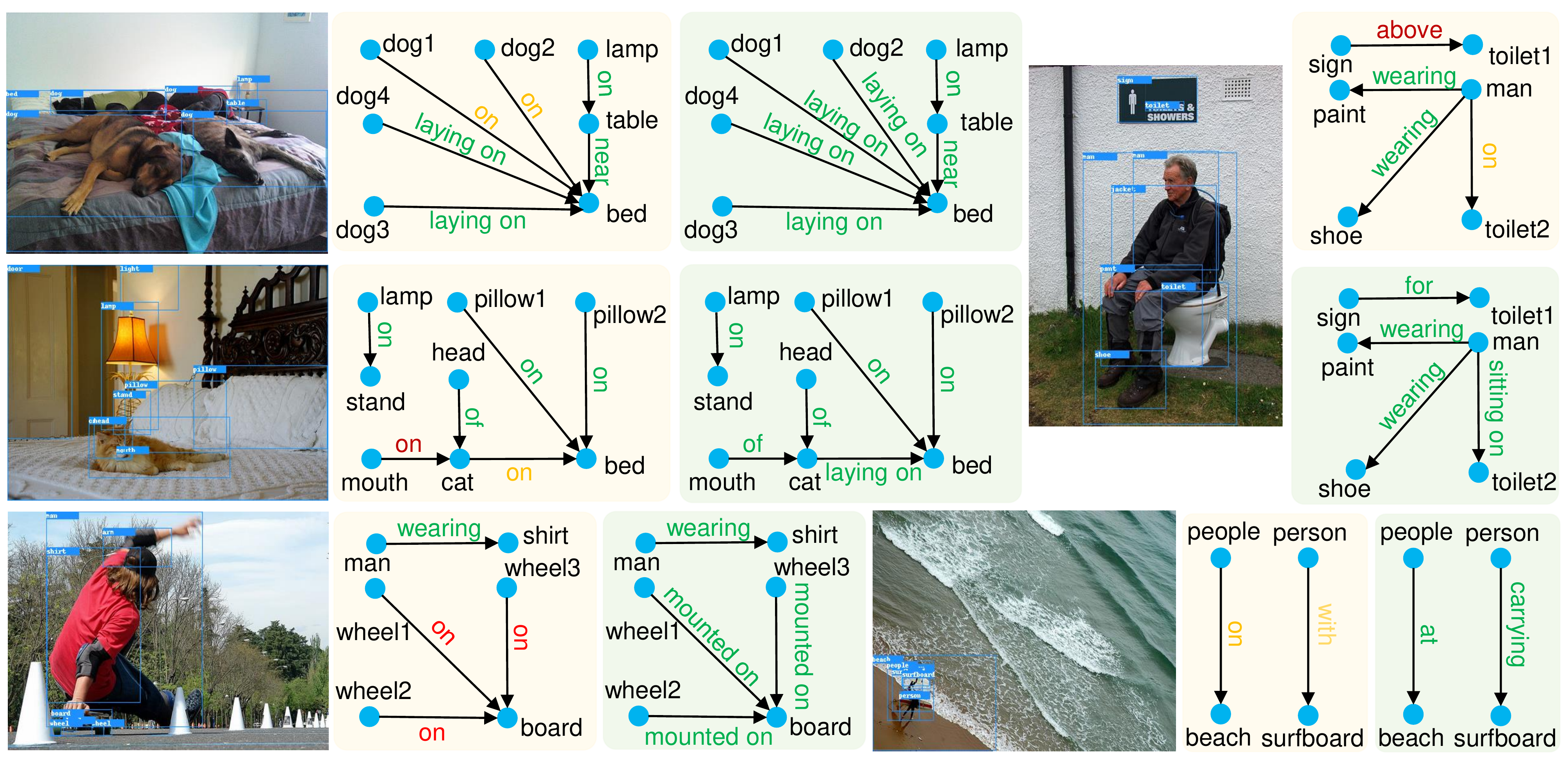}
  \vspace{-1.2em}
  \caption{Results of scene graphs generated from Motifs-baseline (yellow) and Motifs-Ours (green) in PredCls. \textcolor{green}{Green} predicates are correct (\ie, match GT), \textcolor{yellow}{yellow} predicates are acceptable (\ie, does not match GT but still reasonable), and \textcolor{red}{red} predicates are incorrect (\ie, does not match GT and unreasonable).}
  \label{fig:example}
\end{figure*}

\subsubsection{Generated Scene Graphs.}
Some generated scene graphs are visualized in Fig.~\ref{fig:example}, where the scene graphs with a yellow background are generated by Motifs-baseline, and those with a green background are by Motifs-Ours. Only the top-1 prediction and detected boxes overlapped with GT are shown. Comparing the results of the two methods, we can find that our method can detect more informative predicates, such as ``\texttt{laying on}", ``\texttt{sitting on}", and ``\texttt{carrying}". These predicates all have corresponding head predicates that have overlapping or overlaying semantics with them, such as ``\texttt{on}" and ``\texttt{with}". The reason may be that we can reduce the margin of possibility between the head and tail predicates with similar semantics by capturing the correlation of the instance to multiple predicates. Thus, the over-confident predictions on head predicates under long-tailed distribution can be alleviated to some extent. Besides, for triplets that are hard to predict (\eg, ``\texttt{wheel}-\texttt{mounted on}-\texttt{board}" and ``\texttt{sign}-\texttt{for}-\texttt{toilet}"), Motifs-baselines are susceptible to frequency bias to make incorrect predictions (\eg, ``\texttt{wheel}-\texttt{on}-\texttt{board}" and ``\texttt{sign}-\texttt{above}-\texttt{toilet}"), while the predictions of our method are correct. It confirms that reducing the overfitting to the head predicates by training with soft labels is effective.

\subsubsection{Label Semantic Knowledge Distillation Module.}
\begin{figure*}[!t]
  \centering
  \includegraphics[width=0.75\linewidth]{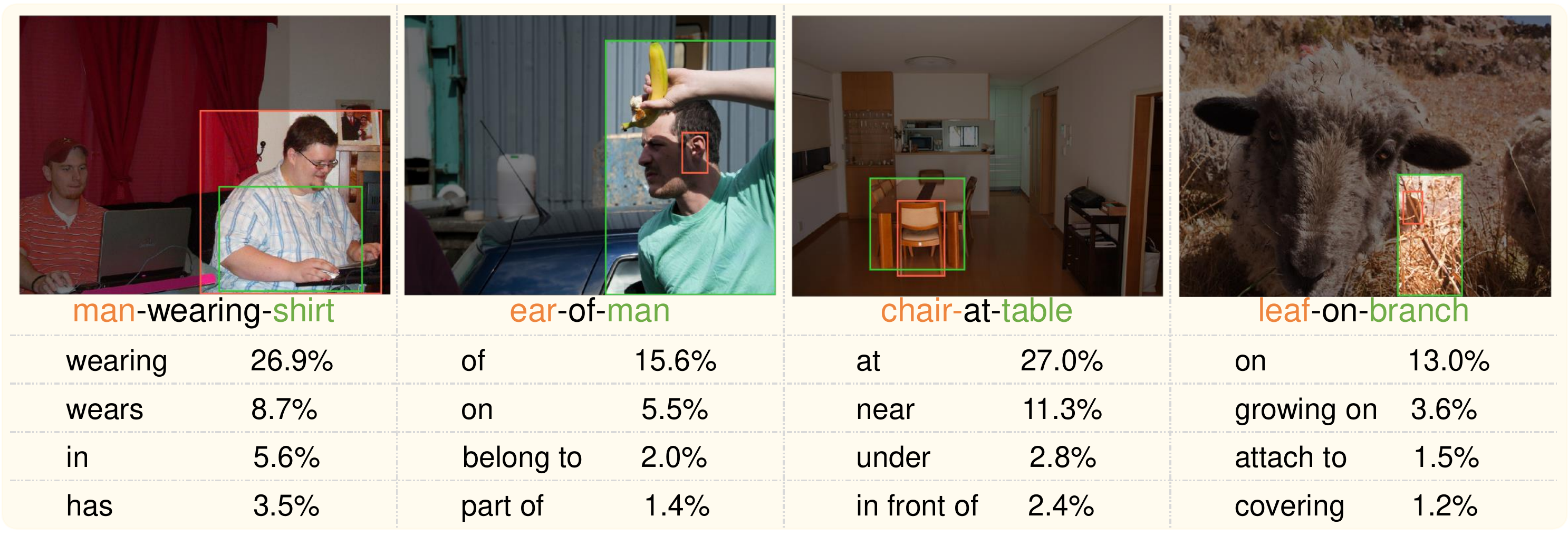}
  \vspace{-0.8em}
  \caption{Results of top 4 predicate categories that are most relevant to the triplet instance without \texttt{background} category.}
  \label{fig:lcm_example}
\end{figure*}

In Fig.~\ref{fig:lcm_example}, we displayed the top 4 predicate categories derived from the iterative self-KD in label semantic knowledge distillation module that are most relevant to each \texttt{subject}-\texttt{predicate}-\texttt{object} triplet instance. From the results, we can observe that these predicate categories associated with a specific instance are all reasonable and even more valuable and informative to the \texttt{subject}-\texttt{object} pair. For example, for the ``\texttt{chair}-\texttt{at}-\texttt{table}" triplet, the two predicate categories ``\texttt{under}" and ``\texttt{in front of}" capture the spatial position relationship between the ``\texttt{chair}" and the ``\texttt{table}" from different angles, and these more exact predicates are more valuable than ``\texttt{at}". In addition, for the ``\texttt{leaf}-\texttt{on}-\texttt{branch}" triplet, predicates ``\texttt{growing on}" and ``\texttt{covering}" represent a dynamic relationship that is more informative than the static ``\texttt{on}". Besides, the predicate category with the max proportion is still GT (\ie, best answer), and the other reasonable tail predicates (\ie, a fair but generally good answer) are assigned proportional probabilities. Thus, the soft labels generated by LS-KD can avoid the over-fitting of the head predicates and promote unbiased prediction to a certain extent. 

\section{Conclusions and Future Work}
In this paper, we point out the drawbacks of the prevalent training paradigm in existing SGG works, \ie, training the predicate classifier with one-hot target labels. To alleviate these drawbacks, we present a novel model-agnostic LS-KD for unbiased SGG. Two different self-KD strategies are designed in LS-KD to predict label semantic distribution (LSD): iterative self-KD and synchronous self-KD. LSD can not only reflect the correlations between subject-object instances and multiple predicate categories, but also generate more robust ``soft" pseudo labels. Extensive results have proved the generality and effectiveness of our method. Moving forward, we are going to: 1) integrate LS-KD into existing debiasing methods whose goal is to solve the long-tailed distribution problem. 2) design better methods to capture correlations among predicate labels.

\section*{Acknowledgments}
This work was supported by the National Key Research \& Development Project of China (2021ZD0110700), the National Natural Science Foundation of China (U19B2043, 61976185), Zhejiang Natural Science Foundation (LR19F020002), Zhejiang Innovation Foundation(2019R52002), and the Fundamental Research Funds for the Central Universities.

\vfill

\end{document}